\documentclass[11pt]{article}

\usepackage[preprint]{acl}

\usepackage{times}
\usepackage{latexsym}

\usepackage[T1]{fontenc}

\usepackage[utf8]{inputenc}

\usepackage{microtype}

\usepackage{inconsolata}

\usepackage{graphicx}
\usepackage{placeins}

\title{Polish Medical Visual Question Answering: Vision-Language Models Underutilize Visual Evidence}

\author{
\textbf{Jakub Pokrywka}\textsuperscript{1}
\quad
\textbf{Łukasz Grzybowski}\textsuperscript{1,2}
\quad
\textbf{Antoni Lasik}\textsuperscript{3}
\\[0.6em]
\textbf{Marek Kubis}\textsuperscript{1}
\quad
\textbf{Jeremi Ignacy Kaczmarek}\textsuperscript{1,4,5}
\quad
\textbf{Wojciech Kusa}\textsuperscript{3}
\\[0.8em]
\textsuperscript{1}Adam Mickiewicz University
\quad
\textsuperscript{2}ARAAI Poland
\quad
\textsuperscript{3}NASK National Research Institute
\\
\textsuperscript{4}Poznań University of Medical Sciences
\quad
\textsuperscript{5}T. Marciniak Lower Silesian Specialist Hospital
}

\begin{document}
\maketitle
\begin{abstract}
We introduce a Polish-language medical visual question answering (VQA)
benchmark, built from Polish Board Certification Examination questions
for licensed physicians and dentists pursuing specialist certification. The
benchmark comprises image-containing questions spanning diverse medical
specialties and visual domains, together with a text-only question answering
(QA) control set. We evaluate Polish-oriented, general-purpose open-weight, and
commercial vision-language models. The task remains challenging: the best model
achieves 79.0\% accuracy on the full VQA set, and only GPT-5.6 surpasses the
approximate human reference on the subset with available candidate responses;
all other evaluated models perform worse than humans. To assess visual grounding,
we compare complete inputs with configurations omitting the image, the question,
or both, and categorize questions by image importance. Models derive more useful
information from the question text than from the image and perform worse on
image-dominant questions. Across both QA and VQA, they nevertheless achieve
above-chance accuracy from the answer choices alone, showing that non-trivial
performance can persist even when key task components are missing.
\end{abstract}

\section{Introduction}

Large language models (LLMs) have recently been evaluated on several Polish medical examination benchmarks, including the Polish Board Certification Examination (pol. \emph{Państwowy Egzamin Specjalizacyjny}, PES), the Medical Final Examination (pol. \emph{Lekarski Egzamin Końcowy}, LEK), the Dental Final Examination (pol. \emph{Lekarsko-Dentystyczny Egzamin Końcowy}, LDEK), and related medical test sets \cite{pokrywka2024gpt4passes297written,grzybowski-etal-2025-polish,lasik2026reassessinghighperformingllmspolish}. These studies showed that modern LLMs can achieve strong results on Polish medical multiple-choice questions and provided evidence on how well models handle specialized medical knowledge in a non-English setting. However, their evaluation protocols were limited to text-only questions. As a result, examination items containing images were excluded, even though visual information is an important component of many real medical tasks and of some PES questions.

This omission leaves an important gap in the evaluation of medical AI systems. While visual question answering (VQA) has been widely studied in English, non-English medical VQA remains relatively underexplored. Polish VQA resources are also limited, especially in specialized domains such as medicine. This is problematic because model performance in English cannot be assumed to transfer directly to Polish, and medical examination questions often require knowledge of domain-specific terminology, clinical conventions, and local examination formats. Consequently, there is a need for benchmarks that evaluate not only medical knowledge in Polish, but also the ability of models to combine Polish clinical text with medical images.

In this work, we evaluate vision-language models (VLMs) on image-containing questions from PES, the Polish Board Certification Examination. These questions are multiple-choice examination items intended for physicians and dentists pursuing specialist certification. They provide a challenging test bed for multimodal medical question answering, as they often require both domain knowledge and interpretation of visual evidence. Importantly, the dataset is not composed solely of classical VQA examples where the image is the central object of a direct visual query. In many cases, the image is only one component of a broader clinical scenario: the question may include a textual patient description, laboratory or diagnostic context, answer choices, and an image such as an electrocardiogram, radiological scan, or clinical photograph. Therefore, the task is better understood as multimodal medical examination question answering rather than simple image recognition or image-centered VQA.

Beyond measuring overall model accuracy, we study how much information models obtain from different parts of the input. Prior work has shown that models can exploit artifacts in multiple-choice answer options or rely disproportionately on textual cues instead of genuinely using visual evidence \cite{mcq-without-questions,mcq-flaws,mirage-medical-vqa-bias}. To examine this issue in the Polish medical examination setting, we evaluate models under controlled input configurations: using only the answer choices, using choices together with the question text, using choices together with the image, and using the full input consisting of choices, question text, and image. This setup allows us to estimate the relative contribution of answer choices, textual context, and visual information.

We also compare performance on image-containing PES questions with performance on a text-only question answering (QA) control set composed of questions that originally did not include images. This comparison allows us to analyze differences between QA- and VQA-style evaluation within the same examination domain. Additionally, we conduct a data contamination analysis to assess whether model performance may have been influenced by prior exposure to the evaluation questions.

Our contributions are as follows:
\begin{itemize}
    \item We create an image-containing question dataset from the Polish Board Certification Examination as a benchmark for Polish medical multimodal question answering.
    \item We evaluate vision-language models on PES questions under several input configurations that separate the effects of answer choices, question text, and images.
    \item We compare model performance on image-containing VQA questions with performance on a text-only QA control set from the same examination domain.
\end{itemize}

\section{Related Work}

\subsection{Polish VQA}

\citet{polish-vqa-dataset} adapt the LLaVA framework to Polish and introduce LLaVA-Bielik and LLaVA-PLLuM. They show that translated and filtered multimodal data can effectively bootstrap Polish VLMs and provide Polish-oriented evaluation resources. reVISION~\cite{revisionciesiolka} evaluates VLMs on Polish multimodal national examination data, extending the exam-based evaluation setting introduced in LLMzSz\L{}~\cite{jassem2025llmzsz} from text-only LLMs to vision-language models. PoVisLE~\cite{anonymous2026povisle} further moves toward Polish-specific vision-language evaluation with emphasis on Polish linguistic and cultural grounding.

Polish is also present in broader multilingual VQA resources. EXAMS-V~\cite{exams-v} includes Polish among multilingual multimodal examination questions, while ~\citet{raj-khan-etal-2021-towards-developing} evaluate cross-lingual transfer to Polish on 500
machine-translated Polish questions.

Our work differs from the aforementioned resources by focusing on specialist-level Polish medical examination questions that require combining clinical text, answer options, and medical images.

\subsection{Medical VQA}

English medical VQA has been studied mainly in radiology, pathology, and biomedical image--text settings. VQA-RAD~\cite{lau2018vqarad} introduced clinically generated questions and answers for radiology images, while the ImageCLEF VQA-Med shared tasks provided a series of radiology-focused medical VQA benchmarks~\cite{ImageCLEFVQAMed2018,ImageCLEFVQAMed2019,ImageCLEFVQAMed2020,ImageCLEFVQAMed2021}. PathVQA~\cite{he2020pathvqa} introduces a new dataset and framework for visual question answering over pathology images. More recent datasets scale medical VQA through visual instruction tuning, including PMC-VQA~\cite{zhang2023pmcvqa} and PubMedVision introduced with HuatuoGPT-Vision~\cite{chen-etal-2024-towards-injecting}.

Non-English and multilingual medical VQA is more limited. SLAKE provides a bilingual English--Chinese medical VQA dataset with semantic labels and medical knowledge~\cite{liu2021slake}. WorldMedQA-V~\cite{matos-etal-2025-worldmedqa} and MMMED~\cite{riccio2025mmmed} evaluate multimodal medical examination questions in multiple languages. Other recent resources address specific languages or domains, including multilingual wound-care VQA~\cite{yim2025woundcarevqa}, Indonesian radiology VQA~\cite{yudhistira2026indorad}, and multilingual hematology VQA~\cite{malik2026wbcmorvqa}.

\subsection{Biases in VQA}

VQA benchmarks often contain linguistic or answer-distribution shortcuts that allow models to answer correctly without sufficient visual grounding. \citet{goyal2017making} addressed this issue by introducing VQA v2, where similar images are paired with the same question but different answers, making the visual signal more important. \citet{agrawal2018dont} further showed that VQA models rely heavily on question-answer priors by introducing VQA-CP, a split with different answer distributions between training and test data. Several works proposed methods to reduce such biases, including adversarial regularization with a question-only model~\cite{ramakrishnan2018overcoming}, RUBi, which downweights examples solvable without the image~\cite{cadene2019rubi}, and visually grounded question encoding~\cite{gouthaman2020reducing}.

This issue is also relevant for modern VLMs and medical VQA. MIRAGE shows that frontier VLMs can generate detailed visual reasoning and obtain high scores on multimodal benchmarks even without image input~\cite{mirage-medical-vqa-bias}. \citet{zhan2023debiasing} propose counterfactual training and a changing-priors medical VQA split to reduce reliance on linguistic shortcuts. Med-BiasX similarly targets medical language biases caused by imbalanced data and question shortcut dependence~\cite{zhu2025medbiasx}. These findings motivate our controlled input configurations, which separately evaluate performance from answer choices, question text, images, and their combination.

\section{Dataset}

\subsection{Examination background}

The dataset used in this work is based on the Polish Board Certification Examination
(PES), a national examination for physicians
and dentists pursuing specialist certification in Poland. Candidates taking PES have
already obtained a medical or dental license and completed the required specialist
training, including clinical practice, courses, internships, and discipline-specific
procedural requirements.

The examination consists of a written and an oral component. The written part is
held separately for each medical or dental specialty and typically contains 120
single-choice questions. Each question has five answer options, exactly one of which
is correct. Most questions are text-only, although some include an accompanying
image. A score of at least 60\% is required to pass the written examination.
Since 2022, candidates who obtain at least 70\% in the written part have been
exempted from the oral examination. Unlike licensing examinations such as LEK and
LDEK, PES questions are not publicly available before the exam, which makes them
a suitable source of challenging specialist-level medical questions.

In this study, we focus on the written part of PES, as it provides standardized
multiple-choice questions with unambiguous correct answers. This format enables
automatic evaluation of model predictions while preserving the specialist-level
medical character of the task.

\subsection{Benchmark construction}

We collected PES examination materials from the Medical Examination Center
(Centrum Egzaminów Medycznych, CEM) website\footnote{\url{https://www.cem.edu.pl/}}, covering examination sessions
from 2023 to 2026. In total, we processed 363 examination sheets. We removed
questions marked by CEM as invalid or no longer aligned with current medical
knowledge.

We then identified examination sheets containing at least one image. This yielded
116 examination sheets with visual material. From these sheets, we extracted all
questions containing images and constructed the VQA subset, consisting of 286
image-containing questions. Each VQA item contains the question text, five answer
choices, the correct answer, metadata describing the examination session and
specialty, and the associated image.

For comparison with text-only question answering, we also constructed a QA control
subset. This subset consists of PES questions that originally did not contain any
images. To make the QA subset comparable to the VQA subset, we selected
specialties for which we had at least 10 image-containing questions. For each
examination sheet in these specialties, we sampled 10 text-only questions. This
procedure resulted in 480 QA questions.

The distribution of VQA and QA questions across specialties is shown in
Table~\ref{tab:question_stats}. The distribution of image-containing questions is
not uniform across medical specialties. Emergency medicine contributes the largest
number of VQA questions, followed by maxillofacial surgery, orthopedics, and
pediatric cardiology. Specialties with fewer than 10 image-containing questions
are grouped into the ``Other specialties'' category. Since the QA subset was
constructed only for specialties with at least 10 VQA questions, no QA items are
assigned to this grouped category.

\begin{table}[t]
\centering
\small
\begin{tabular}{lrr}
\textbf{Specialty} & \textbf{VQA} & \textbf{QA} \\
\hline
Emergency medicine & 62 & 70 \\
Maxillofacial surgery & 37 & 70 \\
Orthopedics & 28 & 70 \\
Cardiology (pediatric) & 28 & 60 \\
Conservative dentistry & 13 & 70 \\
Neurology (pediatric) & 13 & 70 \\
Anesthesiology \& critical care & 10 & 70 \\
Other specialties & 95 & 0 \\
\hline
Total & 286 & 480 \\
\end{tabular}
\caption{Number of VQA and QA questions by medical specialty. Specialties with fewer than 10 VQA questions are grouped as other specialties.}
\label{tab:question_stats}
\end{table}

\subsection{Example PES question}

\begin{figure}[h]
\centering
\includegraphics[width=0.15\textwidth]{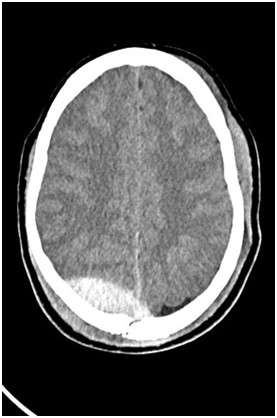}
\caption{\textbf{Image}}
\label{fig:sample_question_ct}
\end{figure}

Figure~\ref{fig:sample_question_ct} presents an example image from an emergency
medicine PES question. The item illustrates the character of the dataset: the model
must combine visual evidence from a head computed tomography (CT) scan with medical knowledge expressed
in the answer options. The correct answer is the false statement about the presented
pathology. The original question is in Polish; the question and answer choices shown
below are an English translation.

\noindent\textbf{Question.}
The attached image shows a head CT scan of a patient after trauma. Indicate the \textbf{false} statement about the presented pathology.

\vspace{0.5em}
\noindent\textbf{Choices:}
\begin{itemize}
    \item[\textbf{A.}] It most commonly results from rupture of the middle meningeal artery.
    \item[\textbf{B.}] In the classic clinical presentation, there is an initial loss of consciousness, followed by a relatively asymptomatic interval, the \textit{lucid interval}.
    \item[\textbf{C.}] In this type of hematoma, blood accumulates between the skull bone and the dura mater.
    \item[\textbf{D.}] A characteristic feature is a lentiform collection of blood that does not cross the cranial sutures to which the dura mater is attached.
    \item[\textbf{E.}] It most commonly results from injury to bridging veins between the surface of the brain and the dural venous sinuses.
\end{itemize}

\vspace{0.5em}
\noindent\textbf{Correct answer:} E

\vspace{0.5em}
\noindent\textbf{Metadata.}
Year: 2023, Quarter: Spring, Specialty: Emergency medicine

\section{Evaluation Methodology}

We evaluate nine open-weight and commercial vision-language models on the
PES-VQA benchmark. The open-weight models comprise three Polish-oriented VLMs:
LLaVA-Bielik-11b-v2.6-instruct, LLaVA-PLLuM-12b-nc-instruct-250715, and
LLaVA-PLLuM-12b-nc-instruct \cite{polish-vqa-dataset}; three Qwen models:
Qwen3.5-397B-A17B, Qwen3.5-9B, and Qwen3.6-27B \cite{qwen3.5}; and
Gemma-4-31B-it \cite{gemma4_31b_it}. The commercial models are
GPT-5.4-nano-2026-03-17 and GPT-5.6-sol \cite{openai2026gpt5api}. Shortened
model labels used in the result tables refer to these full model identifiers.

Among the open-weight models, only the LLaVA variants do not support reasoning;
the others use reasoning by default. In the result tables, the \textbf{R} column
marks reasoning as enabled (\textbf{Y}) or disabled or unavailable (\textbf{N}).
Only the GPT models were evaluated in both configurations: \textit{none}
(\textbf{N}) and \textit{medium} (\textbf{Y}).

\subsection{Input configurations}

The objective of our evaluation extends beyond measuring final accuracy to estimating the amount of information models obtain from different portions of the input.

We therefore evaluate models under several controlled input
configurations. For image-containing questions, we use four settings:
\begin{itemize}
    \item \textbf{C}: answer choices only,
    \item \textbf{C+Q}: answer choices and question text,
    \item \textbf{C+I}: answer choices and image,
    \item \textbf{C+Q+I}: answer choices, question text, and image.
\end{itemize}

The C setting measures whether a model can exploit artifacts, priors, or
statistical regularities in the answer choices without access to the question
itself. The C+Q setting evaluates text-only performance. The C+I setting tests
whether the image provides useful information when the question text is removed.
Finally, C+Q+I corresponds to the complete multimodal examination item as
presented to candidates, whereas the remaining configurations are used only for
our ablation studies.

For the text-only QA control subset, only two configurations are applicable:
C and C+Q.

\subsection{Prompting and evaluation metric}

All experiments were conducted using Polish prompts. The prompt instructed the
model to answer a single-choice medical examination question with options A--E
and to return only a JSON object containing the selected answer. The full prompts,
together with English translations and prompting details, are provided in Appendix~\ref{app:prompts}.

For configurations in which an image was omitted from an image-containing
question, we did not explicitly inform the models that the image was
unavailable. This design choice follows \citet{mirage-medical-vqa-bias}, who
found that model performance declined markedly when models were explicitly
instructed to guess without image access, compared with prompts that implicitly
led them to assume that an image was present.

We report accuracy, defined as the percentage of questions for which the model's
predicted answer matches the official answer key. Since every question has five
answer options and exactly one correct answer, random guessing corresponds to an
expected accuracy of 20\%.

\subsection{Subset with human responses}
Human responses were available for only a subset of the benchmark questions.
We collected these responses and used them to construct a human results subset
for comparing model performance with human performance. The data collection and
alignment procedure is described in Appendix~\ref{app:human_results_subset}.

\subsection{Visual signal categories}

\subsubsection{Image importance}

To characterize how strongly each question depends on its visual material, we
assigned every VQA question to one of three image-importance categories:

\noindent\textbf{0 -- Image non-essential (text sufficient).} The correct
answer can be determined reliably from the question and answer choices without
using the image. The image may illustrate, confirm, or repeat information
already present in the text, but it is not required to solve the question. This
category contains 15 questions (5.2\%).

\noindent\textbf{1 -- Image and text complementary.} Both textual and visual
information are needed to determine the correct answer reliably. The text
provides information independent of the image, such as symptoms, medical
history, test results, or clinical context, which must be combined with the
visual evidence. This category contains 131 questions (45.8\%).

\noindent\textbf{2 -- Image dominant.} The correct answer depends primarily on
interpreting the image. The text contains no substantial clinical information
independent of the image and serves mainly to provide instructions, identify
the type of visual material, or state the task. This category contains 140
questions (49.0\%).
\subsubsection{Visual domains}

We additionally categorized the visual material by content domain. Although
the underlying taxonomy is hierarchical, we report only its top-level
categories:

\noindent\textbf{IMAGE ($n=102$).} Medical images, including radiological and
other diagnostic imaging modalities, microscopy, ophthalmic imaging, and
clinical photography. This category also includes images for which no more
specific subtype was assigned.

\noindent\textbf{WAVEFORM ($n=95$).} Physiological signal traces, including
cardiac, neurophysiological, evoked-potential, and hemodynamic recordings.

\noindent\textbf{PLOT ($n=23$).} Plots presenting measurements, relationships,
or analyses, including audiological, biomechanical, glucose-monitoring,
pressure--volume, radiotherapy, spirometry, and statistical plots.

\noindent\textbf{GRAPHIC ($n=32$).} Explanatory diagrams and schematics
depicting anatomy, biomechanics, classifications, devices, mechanisms, or
procedures.

\noindent\textbf{TABLE ($n=20$).} Visual material organized in tabular form,
including results, comparisons, and matching tables.

\noindent\textbf{COMPOSITE ($n=14$).} Material combining multiple visual or
textual elements, such as device printouts, documents, software screens, or
composite test results.

\noindent\textbf{OTHER ($n=3$).} Visual material that could not be assigned to
any of the categories above.

\section{Data Contamination Analysis}
To detect contamination, we used the Data Contamination Quiz (DCQ) \cite{DCQ} framework. DCQ consists of two stages: quiz creation and examination. First, a frontier LLM generates altered versions of the questions by paraphrasing certain words to break memorization. The tested models are then asked to identify the original question among the paraphrased variants. We used DeepSeek-V4-Pro \cite{deepseekai2026deepseekv4} for quiz creation. Due to computational and cost constraints, we tested one representative model from each evaluated model family for contamination. The results presented in Table \ref{tab:cont} suggest negligible contamination, which should not influence the results of our evaluation. For each model, DCQ reports a closed interval where the lower bound is the bias-corrected minimum contamination level (via Cohen's Kappa) and the upper bound is the maximum raw quiz accuracy across bias-compensated permutations. It is worth noting that the original DCQ prompt explicitly references the dataset name and split as part of the instruction. Since our dataset is not an established, named benchmark, this framing may carry less signal, and the resulting estimates should be interpreted with some caution.
\begin{table}[th]
\centering
\small
\resizebox{\columnwidth}{!}{
\begin{tabular}{lrr}
\hline
\textbf{Model} & \textbf{QA} & \textbf{VQA} \\
\hline
LLaVA-Bielik-11B-v2.6 & (16.86, 27.08) & (15.38, 29.02) \\
LLaVA-PLLuM-12B-250715 & (14.39, 24.38) & (11.19, 22.38) \\
Qwen3.6-27B & (11.27, 11.46) & (10.88, 11.19) \\
Gemma-4-31B-it & (0.03, 16.67) & (0.07, 22.03) \\
GPT-5.4-nano & (17.41, 22.92) & (7.69, 18.53) \\
\hline
\end{tabular}
}

\caption{Contamination level ranges reported using the DCQ methodology on the PES medical QA/VQA datasets (textual part), provided in the format (min contamination, max contamination).}
\label{tab:cont}
\end{table}

\section{Results}

\begin{figure*}[h]
\centering
\includegraphics[width=1\textwidth]{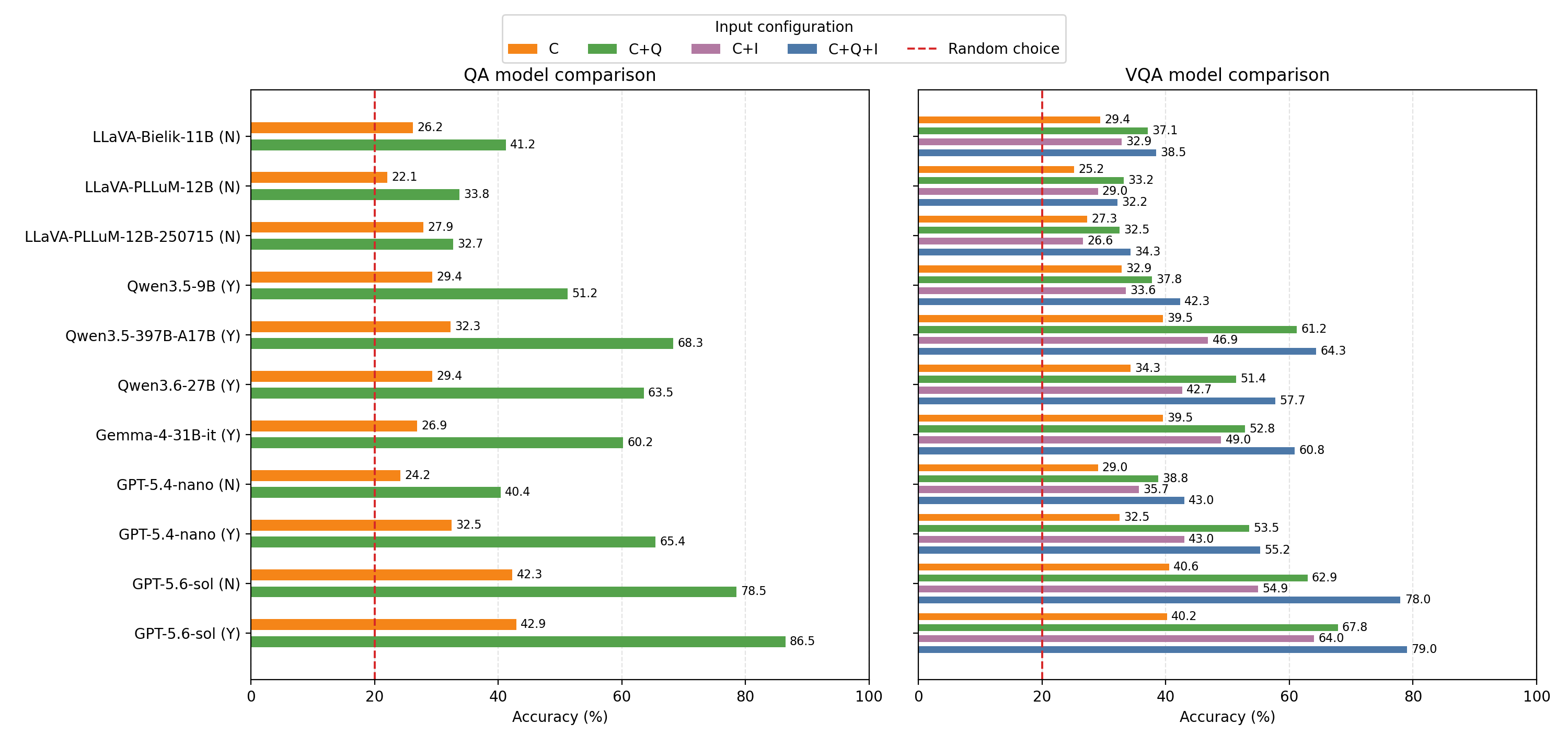}
\caption{Model comparison across input configurations. C+Q represents the complete configuration for QA, while C+Q+I represents the complete configuration for VQA.}
\label{fig:model_comparison}
\end{figure*}

The overall results across input configurations are presented in
Figure~\ref{fig:model_comparison}.

The evaluation conducted with respect to the full dataset shows a strong effect of model size, with larger models achieving higher accuracy.
Among open-weight models small enough to fit on a single consumer-grade GPU, Qwen3.5-9B outperforms all LLaVA-based models.

For the GPT models we evaluated different reasoning-effort settings.
Increasing reasoning effort substantially improves the performance
of GPT-5.4-nano. For GPT-5.6-sol, it improves QA performance but provides only
a small improvement on VQA.

\subsection{Comparison with human examinees}

The comparison on the subset with human responses is presented in
Table~\ref{tab:human_answers}. QA and VQA
have a similar level of difficulty for human examinees, as human accuracy is
nearly identical on the two subsets. For the evaluated models, QA is generally
easier than VQA.

Accuracy above 60\% on both QA and VQA, which corresponds to a passing score
for human examinees, is achieved by Qwen3.5-397B-A17B, Gemma-4-31B-it, and
GPT-5.6. Among the evaluated models, only GPT-5.6-sol outperforms human
examinees on average, under both the \textit{none} and \textit{medium}
reasoning-effort settings. This finding indicates that the dataset is
particularly challenging for models. The commercial GPT-5.6 model also
substantially outperforms the open-source models.

\subsection{Model performance with incomplete inputs}

We analyze model performance under incomplete-input scenarios separately for QA
and VQA.

\begin{table}[t]
\centering
\small
\setlength{\tabcolsep}{4pt}
\begin{tabular}{@{}llcc@{}}
\hline
\textbf{Model} & \textbf{R} & \textbf{QA} & \textbf{VQA} \\
\hline
LLaVA-Bielik-11B       & N & 43.35 & 39.68 \\
LLaVA-PLLuM-12B        & N & 34.84 & 33.33 \\
LLaVA-PLLuM-12B-250715   & N & 34.04 & 35.32 \\
Qwen3.5-9B             & Y & 52.39 & 41.27 \\
Qwen3.5-397B-A17B      & Y & 68.88 & 63.10 \\
Qwen3.6-27B            & Y & 64.10 & 57.94 \\
Gemma-4-31B-it         & Y & 60.37 & 61.51 \\
GPT-5.4-nano           & N & 41.22 & 41.67 \\
GPT-5.4-nano           & Y & 64.36 & 54.76 \\
GPT-5.6-sol                & N & 78.99 & 76.59 \\
GPT-5.6-sol                & Y & 86.70 & 77.78 \\
\hline
Human examinees        &   & 69.75 & 70.14 \\
\hline
\noalign{\vskip 2pt}
\multicolumn{4}{@{}c@{}}{\textit{Subset statistics}} \\
Examinees &  & 2,103  & 2,001 \\
Exams     &  & 47     & 55 \\
Questions &  & 376    & 252 \\
Answers   &  & 11,167 & 6,614 \\
\hline
\end{tabular}
\caption{Model and human accuracy on the QA and VQA subsets, together with
subset statistics. Human accuracy
is the percentage of examinee answers matching the answer key.}
\label{tab:human_answers}
\end{table}

\subsubsection{QA}
All models perform above the random-guessing baseline of 20\% on QA when given
only the answer choices, without the question (C configuration). One possible explanation is that
the intended question can sometimes be inferred from the answer choices, for
example when the task is to identify a true statement. Alternatively, some
choices may contain an intrinsic error, such as an incorrect justification,
that can be detected without knowing the question. The best-performing model
configuration, GPT-5.6-sol, selects the correct answer in 42.9\% of cases
without access to the question.

\subsubsection{VQA}
In the choices-only configuration (C), performance is similar to that observed
on QA. When the input is incomplete and either the question or the image is
missing (C+I or C+Q), the models perform better than with the answer choices
alone (C), but worse than with the complete input (C+Q+I). Moreover, removing
the image (C+Q) is less detrimental than removing the question (C+I), which
shows that the models make greater use of the question text. This pattern holds
for every model except LLaVA-PLLuM, which performs relatively poorly overall.
These results show that the models perform well under incomplete-information
conditions, although their ability to identify the correct answer decreases as
more information is removed.

\subsection{Model performance by visual category}

\subsubsection{Image importance}

Table~\ref{tab:vqa_accuracy_by_image_importance} reports accuracy for the
pooled text-sufficient and complementary categories (0+1) and for the
image-dominant category (2). Every evaluated model performs worse on
image-dominant questions under both C+Q and C+Q+I. The difference under C+Q is
expected because this configuration omits the image, which category~2 questions
primarily require. More notably, the same ordering persists with the complete
C+Q+I input. This pattern is consistent with the
input-ablation results indicating that the models rely more heavily on textual
cues than on visual evidence.

\begin{table*}[t]
\centering
\small
\setlength{\tabcolsep}{4pt}
\begin{tabular}{@{}ll|rr|rr@{}}
\hline
\textbf{Model} & \textbf{R} & \multicolumn{2}{c|}{\textbf{C+Q}} & \multicolumn{2}{c}{\textbf{C+Q+I}} \\
\multicolumn{2}{@{}l|}{\textbf{Image importance category}} & \textbf{0+1 ($n=146$)} & \textbf{2 ($n=140$)} & \textbf{0+1 ($n=146$)} & \textbf{2 ($n=140$)} \\
\hline
LLaVA-Bielik-11B & N & 44.52 & 29.29 & 47.95 & 28.57 \\
Qwen3.5-397B-A17B & Y & 76.71 & 45.00 & 76.03 & 52.14 \\
Gemma-4-31B-it & Y & 68.49 & 36.43 & 73.97 & 47.14 \\
GPT-5.4-nano & N & 47.26 & 30.00 & 52.05 & 33.57 \\
GPT-5.4-nano & Y & 67.81 & 38.57 & 64.38 & 45.71 \\
GPT-5.6-sol & N & 82.19 & 42.86 & 83.56 & 72.14 \\
GPT-5.6-sol & Y & 80.82 & 54.29 & 84.25 & 73.57 \\
\hline
\end{tabular}
\caption{VQA accuracy by image-importance category for the C+Q and C+Q+I input configurations. Categories 0 and 1 are pooled; category 2 contains image-dominant questions.}
\label{tab:vqa_accuracy_by_image_importance}
\end{table*}

\subsubsection{Visual domains}

Table~\ref{tab:vqa_accuracy_by_visual_domain} shows accuracy across the seven
top-level visual domains. WAVEFORM tends to be among the best-performing
well-represented domains. These comparisons should be interpreted cautiously because the domain sizes
are unequal and the relative performance patterns across domains differ between
models.

\begin{table*}[t]
\centering
\small
\setlength{\tabcolsep}{3pt}
\begin{tabular}{@{}ll|rrrrrrr@{}}
\hline
\textbf{Model} & \textbf{R} & \textbf{IMAGE} & \textbf{WAVEFORM} & \textbf{PLOT} & \textbf{GRAPHIC} & \textbf{TABLE} & \textbf{COMPOSITE} & \textbf{OTHER} \\
\multicolumn{2}{@{}l|}{\textbf{Domain size}} & $n=102$ & $n=95$ & $n=23$ & $n=32$ & $n=20$ & $n=14$ & $n=3$ \\
\hline
LLaVA-Bielik-11B & N & 35.29 & 50.53 & 34.78 & 28.12 & 25.00 & 42.86 & 0.00 \\
Qwen3.5-397B-A17B & Y & 59.80 & 71.58 & 69.57 & 62.50 & 60.00 & 57.14 & 66.67 \\
Gemma-4-31B-it & Y & 58.82 & 71.58 & 69.57 & 34.38 & 70.00 & 35.71 & 66.67 \\
GPT-5.4-nano & N & 40.20 & 54.74 & 43.48 & 31.25 & 30.00 & 28.57 & 33.33 \\
GPT-5.4-nano & Y & 56.86 & 62.11 & 60.87 & 37.50 & 55.00 & 14.29 & 100.00 \\
GPT-5.6-sol & N & 80.39 & 83.16 & 69.57 & 62.50 & 75.00 & 78.57 & 100.00 \\
GPT-5.6-sol & Y & 76.47 & 82.11 & 78.26 & 71.88 & 85.00 & 85.71 & 100.00 \\
\hline
\end{tabular}
\caption{VQA accuracy by visual domain. Questions containing panels from multiple domains contribute to each corresponding domain.}
\label{tab:vqa_accuracy_by_visual_domain}
\end{table*}

\section{Conclusion}

In this work, we introduce the first Polish-language medical VQA benchmark,
accompanied by a text-only QA control subset. Human examinees achieve nearly
identical accuracy on the two subsets, indicating that the tasks are comparable
in difficulty.

We evaluate Polish-oriented and general-purpose open-weight models, as well as
proprietary commercial systems. Although QA and VQA are similarly difficult for
human examinees, most evaluated models perform worse on VQA. The input-ablation
experiments further show that models derive more useful information from the
question text than from the image: on VQA, they generally perform better without
the image (C+Q) than without the question text (C+I). Moreover, accuracy is lower
on image-dominant questions than on questions for which textual information is
sufficient or complementary, even when the models receive the complete
multimodal input. Together, these findings suggest that current models rely more
heavily on textual cues than on visual evidence when answering Polish medical
examination questions.

Across both QA and VQA, the models also achieve above-chance accuracy when given
only the answer choices, suggesting that the options provide textual cues that
support inference even when both the question and the image are unavailable.
This result complements the image-free evaluation of
\citet{mirage-medical-vqa-bias}, who show that VLMs can achieve high scores on
multimodal benchmarks without visual input and identify textual cues as a source
of non-visual inference. Our more restrictive choices-only setting further shows
that some usable signal may reside in the answer options themselves. Together,
these findings call for caution when interpreting benchmark performance: high
accuracy does not necessarily reflect robust medical or multimodal competence
when a task remains partially solvable from incomplete inputs. This
consideration is especially important in medical settings, where model outputs
may have significant consequences.

\FloatBarrier
\section*{Limitations}

Our evaluation covers a selected set of models rather than the full space of
currently available vision-language models. The number of open-weight and
commercial systems is growing rapidly, making an exhaustive comparison
impractical. We therefore focused on models that we consider representative of
different relevant categories, including Polish-oriented VLMs, general
multilingual VLMs, and commercial systems.

We did not evaluate all possible inference settings. In particular, we did not
systematically test all reasoning-effort levels, decoding configurations, or
image-resolution variants. These factors may affect model performance, especially
for visually demanding medical questions. Our results should therefore be
interpreted as performance under the default prompting and inference setup used
in this work, rather than as the maximum attainable performance of each model.

We did not include vision-language models specifically trained for medicine.
Previous work on Polish medical examination benchmarks has shown that general
purpose LLMs can outperform models adapted to the medical domain
\cite{grzybowski-etal-2025-polish}. Nevertheless, this observation comes mainly
from text-only evaluation and may not fully transfer to multimodal medical tasks.

The dataset is limited in size and specialty coverage. The VQA subset contains
286 image-containing questions, and the distribution across specialties is uneven.
Some specialties are represented by many more items than others, while specialties
with few image-containing questions provide only limited evidence about model
performance. At the same time, the questions are high-quality examination items
prepared for specialist medical certification, which makes them a valuable
resource despite their limited number.

The benchmark reflects the structure of Polish board certification examinations
rather than the full range of clinical practice. The questions test specialist
medical knowledge under a standardized written-exam format, but they do not
capture interactive patient assessment, longitudinal decision-making, procedural
skills, communication, or responsibility for real-world outcomes.

\section*{Ethical Considerations}

The questions used in this study originate from materials published by the Polish
Medical Examination Center (Centrum Egzaminów Medycznych, CEM). We did not
author the original examination questions; our contribution consists of collecting,
processing, structuring, and evaluating them as a benchmark for multimodal
medical question answering. We preserve the examination character of the items
and use them only for research evaluation.

Performance on written medical examinations captures only a limited part of
medical competence. Becoming a licensed physician or dentist in Poland requires
extensive education, supervised clinical training, practical experience, and
formal certification. A model that performs well on multiple-choice exam
questions should therefore not be described as equivalent to a clinician, nor
should such results be used to claim that models can replace medical
professionals.

This limitation is particularly important for multimodal medical questions.
Clinical work requires gathering information from patients, performing physical
examinations, interpreting diagnostic tests in context, weighing
contraindications and comorbidities, communicating uncertainty, and making
decisions under incomplete information. A benchmark based on static examination
items cannot evaluate these abilities comprehensively.

LLMs and VLMs may be useful in medical education, information retrieval, and
decision-support workflows, but they can also generate incorrect, incomplete, or
misleading outputs. In medical settings, such errors may create substantial risks
if model responses are treated as authoritative. Any practical deployment of
these systems should therefore include oversight by qualified healthcare
professionals, clear communication of model limitations, and compliance with
applicable ethical, clinical, and regulatory standards.

\bibliography{custom}

\appendix

\section{Prompts and Output Format}
\label{app:prompts}

We used only two versions of the system prompt. The first version was used when
the question text was available, i.e., in the C+Q and C+Q+I configurations. The
second version was used when the question text was removed, i.e., in the C and
C+I configurations. The presence of an image did not change the textual prompt:
in image-based configurations, the image was simply attached to the model input
together with the same textual prompt used in the corresponding non-image
configuration.

All experiments were conducted using Polish prompts. For readability, we also
provide English translations below. The English versions were not used
as separate experimental prompts; they are included only as translations of the
Polish prompts.

\subsection{System prompt with question text}

This prompt was used for configurations that included the question text, i.e.,
C+Q and C+Q+I.

\paragraph{Polish prompt.}
\begin{quote}
\small
Odpowiadasz na pytania jednokrotnego wyboru A--E z testu medycznego dla lekarzy.

Wybierz dokładnie jedną odpowiedź spośród: A, B, C, D, E.

Zwróć wyłącznie poprawny obiekt JSON w następującej strukturze:

\texttt{\{"response": "<LETTER>"\}}

Zastąp \texttt{<LETTER>} jedną wybraną literą: A, B, C, D albo E.
Pole \texttt{"response"} musi być typu string.

Bez wyjaśnień, bez komentarzy, bez dodatkowego tekstu.
\end{quote}

\paragraph{English translation.}
\begin{quote}
\small
You answer single-choice questions with options A--E from a medical examination
for physicians.

Select exactly one answer from: A, B, C, D, or E.

Return only a valid JSON object with the following structure:

\texttt{\{"response": "<LETTER>"\}}

Replace \texttt{<LETTER>} with one selected letter: A, B, C, D, or E.
The \texttt{"response"} field must be a string.

No explanations, comments, or additional text.
\end{quote}

\subsection{System prompt without question text}

This prompt was used for configurations without the question text, i.e., C and
C+I. In these settings, the model received only the answer options and,
depending on the configuration, optionally the image.

\paragraph{Polish prompt.}
\begin{quote}
\small
Odpowiadasz na pytania jednokrotnego wyboru A--E z testu medycznego dla lekarzy.
Nie otrzymujesz treści pytania. Masz tylko odpowiedzi A--E.

Mimo braku treści pytania spróbuj wskazać najbardziej prawdopodobną poprawną
odpowiedź.
Wybierz dokładnie jedną odpowiedź spośród: A, B, C, D, E.

Zwróć wyłącznie poprawny obiekt JSON w następującej strukturze:

\texttt{\{"response": "<LETTER>"\}}

Zastąp \texttt{<LETTER>} jedną wybraną literą: A, B, C, D albo E.
Pole \texttt{"response"} musi być typu string.

Bez wyjaśnień, bez komentarzy, bez dodatkowego tekstu.
\end{quote}

\paragraph{English translation.}
\begin{quote}
\small
You answer single-choice questions with options A--E from a medical examination
for physicians.
You do not receive the question text. You only have the answer options A--E.

Despite the absence of the question text, try to identify the most likely
correct answer.
Select exactly one answer from: A, B, C, D, or E.

Return only a valid JSON object with the following structure:

\texttt{\{"response": "<LETTER>"\}}

Replace \texttt{<LETTER>} with one selected letter: A, B, C, D, or E.
The \texttt{"response"} field must be a string.

No explanations, comments, or additional text.
\end{quote}
\subsection{Output schema}

For models supporting structured outputs, the response was constrained to a JSON
object containing exactly one required field, \texttt{response}. This field was
required to be a string and could take only one of five values corresponding to
the answer options: A, B, C, D, or E. No additional fields were allowed.

\section{Subset with Human Responses}
\label{app:human_results_subset}

To place model performance in the context of human performance, we additionally
collected anonymized candidate answers published by CEM. For each PES session,
CEM publishes anonymized answer sheets of individual examinees, listing the
option selected for each question, together with the official answer key.
Linking these answers to our benchmark is not straightforward. Our analysis of
the published materials revealed that each examination exists in two versions
that differ in question numbering, while only one version of the examination
sheet is published. As a consequence, a given question number in the answer
statistics does not necessarily correspond to the same question in the
published sheet.

To ensure correct alignment, we merged the candidate answers with our question
set using the official correct answer as a consistency check. For every matched
item, the expected correct answer in the published sheet had to agree with the
correct answer reported in the answer statistics. Since a single question can
match by chance, we further restricted the human answers subset to examinations
for which more than one question from our benchmark was available, and required
the expected correct answers to match for all available questions from that
examination. We excluded examinations that did not satisfy this condition, as
we could not reliably determine which version they corresponded to.

This procedure yielded a subset of questions with associated human answers, for
which we computed the accuracy of examinees. Since this subset is smaller than
the full benchmark, we treated the resulting value as an approximate human
reference point rather than an item-level comparison with model accuracy.

\section{Usage of GenAI in Research}

We used ChatGPT and Codex to assist with manuscript writing, code development,
and literature searches. All generated content and outputs were reviewed and
verified by the authors, who remain fully responsible for the final manuscript.

\end{document}